\documentclass[10pt,twocolumn,letterpaper]{article}

\usepackage{cvpr}
\usepackage{times}
\usepackage{epsfig}
\usepackage{graphicx}
\usepackage{amsmath}
\usepackage{amssymb}
\usepackage[american]{babel}

\usepackage[breaklinks=true,bookmarks=false]{hyperref}

\cvprfinalcopy 

\def\cvprPaperID{****} 

\begin{document}

\title{Temporal Modeling Approaches for Large-scale\\ Youtube-8M Video Understanding}

\author{Fu Li, Chuang Gan, Xiao Liu, Yunlong Bian, Xiang Long, Yandong Li, Zhichao Li, Jie Zhou, Shilei Wen\\
Baidu IDL \& Tsinghua University\\
}

\maketitle

\begin{abstract}
  This paper describes our solution for the video recognition task of the Google Cloud \& YouTube-8M Video Understanding Challenge that ranked the 3rd place. Because the challenge provides pre-extracted visual and audio features instead of the raw videos, we mainly investigate various temporal modeling approaches to aggregate the frame-level features for multi-label video recognition. Our system contains three major components: two-stream sequence model, fast-forward sequence model and temporal residual neural networks. Experiment results on the challenging Youtube-8M dataset demonstrate that our proposed temporal modeling approaches can significantly improve existing temporal modeling approaches in the large-scale video recognition tasks. To be noted, our fast-forward LSTM with a depth of 7 layers achieves 82.75\% in term of GAP@20 on the Kaggle Public test set.
\end{abstract}

\section{Introduction}
Video understanding is a challenging task which has received significant research attention in computer vision and machine learning. The ubiquitous video capture devices have created videos far surpassing what we can watch. Therefore, it has been a pressing need to develop automatic video understanding algorithms for various applications.

To recognize actions and events in videos, existing approaches based on deep convolutional neural networks (CNNs)~\cite{Sports1M,Twostream,devnet,C3D} and/or recurrent networks~\cite{LSTM,ULSTM,hochreiter1997long,cho2014properties} have achieved state-of-the-art results. However, due to the lack of publicly available datasets, existing video recognition approaches are restricted to small-scale data, while large-scale video understanding remains an under-addressed problem.  To remedy this issue, Google releases a new web crawled large-scale video dataset, named as YouTube-8M, which contains over 7 million YouTube videos with a vocabulary of 4716 classes. A video may have multiple tag classes and the average number of tag classes per video is 1.8. Prior to this, Gan \textit{et. al}~\cite{gan2016webly,gan2016you} also investigated to learn video recognition models using Web videos and images.

Another appealing point of the Youtube-8M dataset is that this competition only provides the pre-extracted visual and audio features from every second of video instead of raw videos.  We can neither train different CNNs architectures nor learn as optical flow features from the raw videos. Therefore, we focus on temporal modeling approaches to aggregate the frame-level features that yield robust and discriminative video representation for further multi-label recognition. Particularly, we propose three novel temporal modeling approaches, namely two-stream sequence model, fast-forward sequence model and temporal residual neural networks. Experiment results verity the effectiveness of the three models over the traditional temporal modeling approaches. We also find that these three temporal modeling approaches are complementary with each others and lead to the state-of-the-arts performances after ensemble.

The remaining sections are organized as follows. Section \ref{sec:2} presents our temporal modeling approach to learn robust and discriminative video feature representation for recognition. Section~\ref{sec:3} reports empirical results, followed by discussion and conclusion in Section \ref{sec:4}.

\section{Approach}
\label{sec:2}

In this section, we describe our three families of temporal approaches respectively.

\begin{figure*}
   \centering
   \includegraphics[width=1\linewidth]{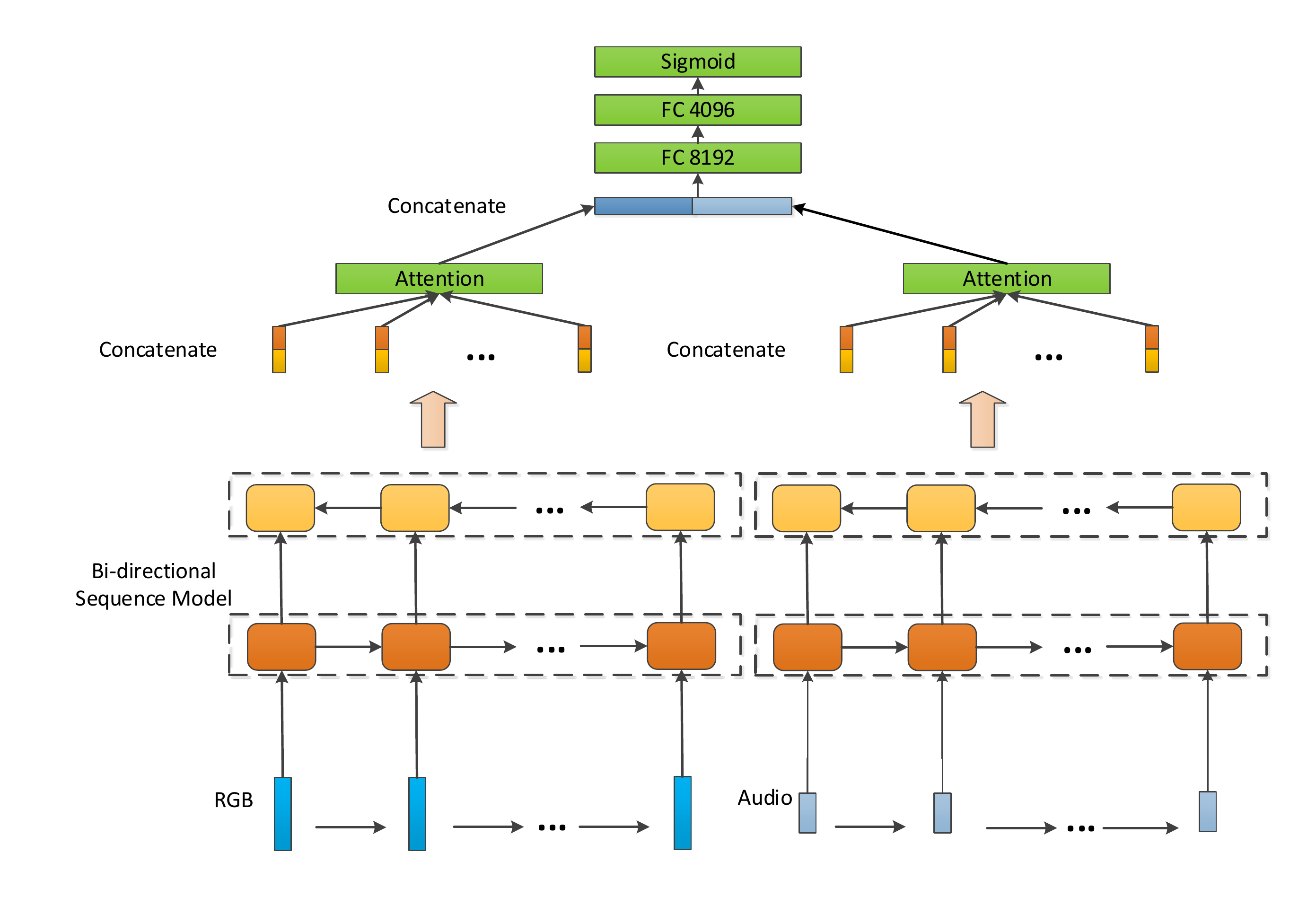}
   \caption{The architecture of our proposed two-stream LSTM model.}
   \label{fig:two_stream}
\end{figure*}

\subsection{Two-stream Sequence Models}
Our two stream sequence models build upon the bidirectional LSTM~\cite{hochreiter1997long} and GRU~\cite{cho2014properties}, since they have shown strong temporal modeling abilities for video recognition. The challenge here is how to incorporate the visual and audio information contained in the videos. In order to best take the advantage of multi-modal clues, we propose several sequence architectures to fuse these two modality features.

The original two-stream CNN~\cite{Twostream} framework trains CNNs with RGB and optical flow features separately, and then relies on a late score fusion strategy to leverage the complementary nature of the two modalities. Recently, Ma \textit{et. al}~\cite{ts_lstm} has proposed a temporal segment RNN network by first concatenating the two modality features and then fed them into one LSTM to achieve video recognition.

Different from them, we propose to train two bidirectional LSTMs or GRUs models (i.e. one for RGB features, and the other for audio features). Attention layers are inserted after the sequence models and attended feature vectors from two modalities are then concatenated. Finally, the concatenated feature vector is fed into two fully-connected layer and a sigmoid layer sequentially for multi-label classification. We outline the framework in Figure~\ref{fig:two_stream}. Experiments results verity the effectiveness of the our proposed two-stream sequence model approaches over other alternative two-stream fusion approaches.

\subsection{Fast-forward Sequence Models}
Recently, we have witnessed the success of deep CNNs on large-scale image classification~\cite{VeryDeep,go_deeper,resnet}. Typically, models
with deeper convolution layers outperform shallow ones. However, this success has not been transferred to the sequence models that used in video recognition tasks. The best sequence models reported in literature are still shallow models. The phenomenon is caused by two reasons. First, it is impossible to explore deeper sequence models in the pre-existing small-scale video recognition dataset~\cite{UCF101,HMDB}, which only contain around 10 thousands videos. Second, the optimization of deeper sequence model is much more challenging than training deeper CNNs because the existence of many more nonlinear activations and the recurrent computation results in smaller and instable gradient.

The new Youtube8M dataset sheds light on opportunities to explore sequence models with deep architectures. Since large-scale video recognition is a very difficult and challenging problem, we believe that deeper sequence models with more complex architecture is necessary for capturing the temporal relationship between frames. In the competition, we focus on enhancing the complexity of the sequence model by increasing the model depth. However, we observe that naively increasing the depth of the LSTM and GRU still entails to overfitting and optimization difficulties, and thus always have negative results for the video recognition. This phenomenon is consistent with the results reported by the original Youtube8M technique report~\cite{YouTube8M}.
\begin{figure*}
   \centering
   \includegraphics[width=1\linewidth]{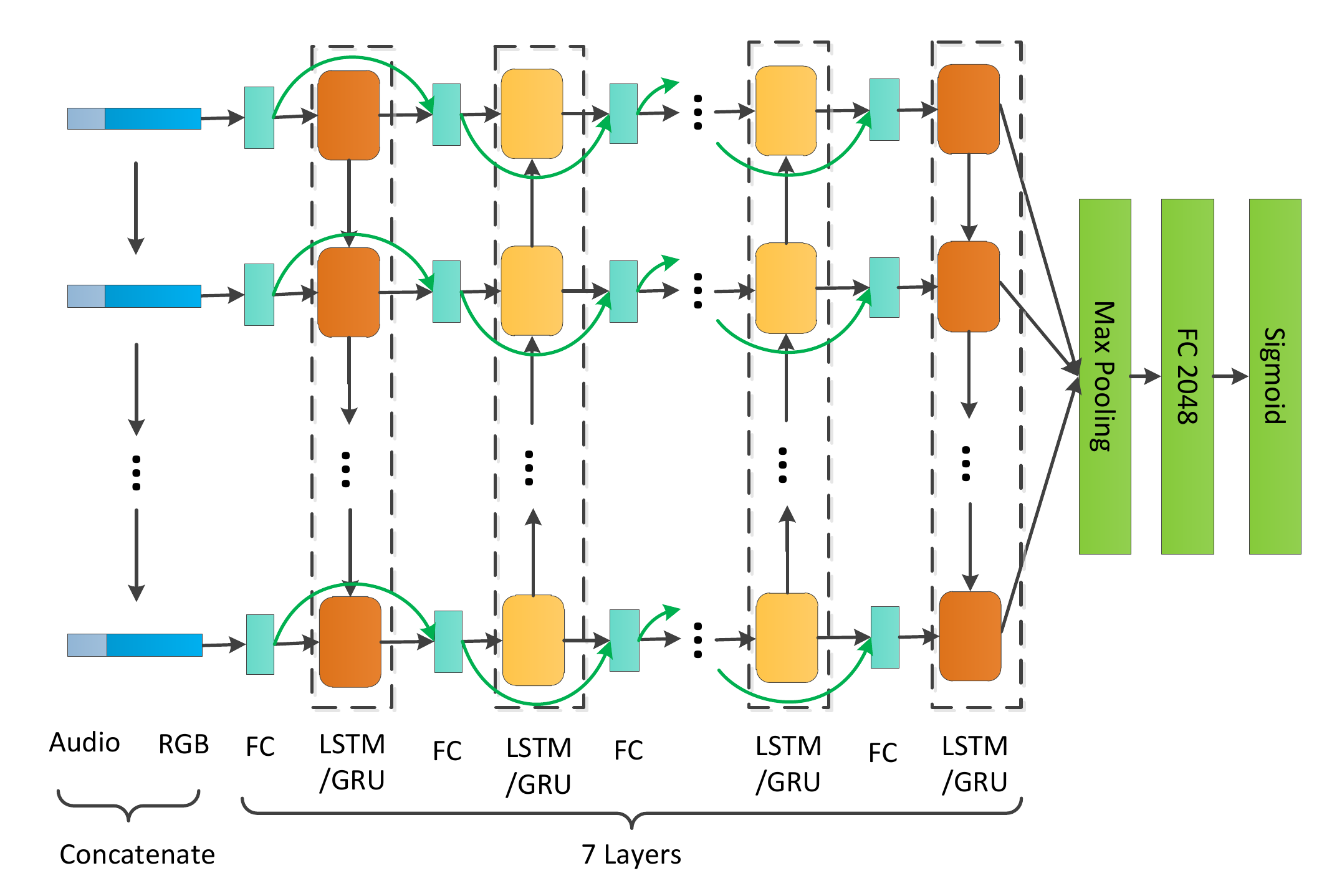}
   \caption{The architecture of our proposed fast-forward sequence models.}
   \label{fig:fast_forward}
\end{figure*}

To address these challenges, we explore a novel deep LSTM/GRU architecture by adding the fast-forward connections~\cite{zhou2016deep} to sequence models, which plays an essential role in building a sequence model with 7 bidirectional LSTMs. We outline the framework in the figure~\ref{fig:fast_forward}. We first concatenate the RGB and audio features of each frame together and then fed them into the fast-forward sequence model. The fast-forward connections are added between two feed-forward computation blocks of adjacent recurrent layers. Each fast-forward connection takes the outputs of previous fast-forward and recurrent layer as input, and use a fully-connected layer to embed them. The fast-forward connect provides a fast path for information to propagate, so we call the path fast-forward connections. We will introduce more detail of our proposed fast-forward sequence model and implementation details in a following technique report.

\subsection{Temporal Residual Neural Networks}
Although the power of recurrent models (LSTMs and GRUs) have been widely acknowledged, recent sequential convolution architectures \cite{deep_speaker, ts_lstm} show strong potentials for various temporal modeling tasks. Li \textit{et. al} ~\cite{deep_speaker} proposed a temporal ResCNN based neural speaker recognition system for speaker identification and verification. Ma \textit{et. al}~\cite{ts_lstm} proposed a temporal-inception architecture for video recognition, and achieved state-of-the-art results on UCF101 and HMDB51 datasets.

In the competition, we investigate the usage of temporal convolution neural networks for temporal modeling on video recognition. In contrast with \cite{ts_lstm} that performs convolutions on frame-level features to learn global video-level representations, we combine convolution and recurrent neural networks to take the advantages of both models. The temporal convolution neural networks are utilized to transform the original frame-level features into a more discriminative feature sequence, and LSTMs are used for final classification.
\begin{figure*}
   \centering
   \includegraphics[width=1\linewidth]{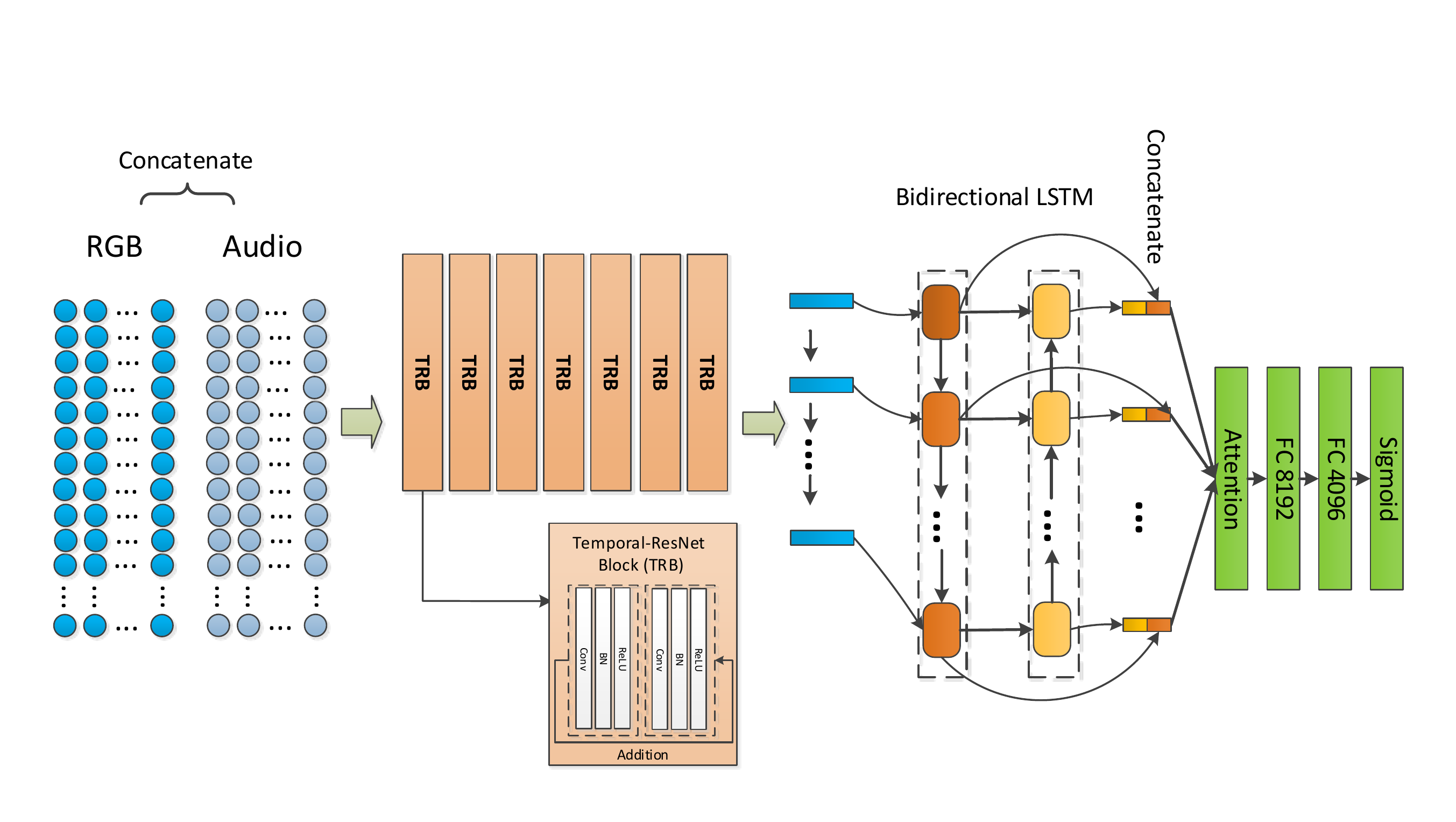}
   \caption{The architecture of our proposed temporal residual CNNs.}
   \label{fig:cnn}
\end{figure*}

The architecture of the proposed Temporal CNN is illustrated in Figure~\ref{fig:cnn}. RGB and audio features in each frame are concatenated and zero-valued features are padded to make fixed length data.
The size of the resulted input data is $ 4000\times 1152 \times 300 $, where 4000, 1152, and 300 indicates mini-batch size, channel number, and length of frames, repsectively. We then propagate the batch data into a Temporal Resnet, which is a stack of 9 Temporal Resnet Blocks (TRB), and each TRB consists of two temporal convolutional layers (followed by batch norm and activation) and a shortcut connection. We use 1024 $3 \times 1$ filters for all the temporal convolution layers. The output of the temporal CNN is then fed into a bidirectional LSTM with attention.

\section{Experiment}
\label{sec:3}
In this section, we present the dataset, experiment setting and our experimental results.

\subsection{Dataset}

We conduct experiment on the challenging Youtube-8M dataset~\cite{YouTube8M}.
This dataset contains around 7 million Youtube videos. Each video is annotated with one or multiple tags.
In the competition,  visual and audio features are pre-extracted and provided with the dataset for each second of the video.
Visual features are obtained by the Google Inception CNN pre-trained on the ImageNet~\cite{ImageNet}, followed by the PCA-compression into a 1024 dimensional vector. The audio features are extracted from a pre-trained VGG~\cite{VeryDeep} network.
In the official split, the dataset is divided into three parts: 70\% for training, 20\% for validation, and 10\% for testing.
In practice, we only maintain 60K videos from the official validation set to cross validate the parameters. Other videos in the validation set are included into the training set. We observe that this strategy can slightly improve the classification performances. Results are evaluated using the Global Average Precision (GAP) metric at top 20 as used in the Youtube-8M Kaggle competition.

\begin{table}[t]
    \centering
 \caption{Comparison results on Youtube8M test set.}
\label{tab:result}

    \begin{tabular}{c|c}

     Method &   GAP@20  \\

     \hline

    Video-level & 0.80824\\
    VLAD & 0.80423\\
    Temporal CNN & 0.80889\\
    Two-stream LSTM  &0.82172\\
    Two-stream GRU &0.82366\\
   \hline
    Fast-forward LSTM & 0.81885\\
    Fast-forward GRU  &0.81970\\

    Fast-forward LSTM (depth7) &\textbf{0.82750}\\

       Ensemble  & \textbf{0.84542} \\
   \hline

\end{tabular}

\end{table}

\subsection{Experiment Results}

Table~\ref{tab:result} reports the performance of individual models on the Youtube8M test set. For the video-level approach, we use the average pooling to aggregate the frame-level feature vector. For VLAD encoding based approaches, we use 256 cluster centers followed by signed square root and L2 normalizations as suggested in~\cite{arandjelovic2013all,xu2015discriminative}. We then fed these representations into a MLP classifier to obtain the final video classification scores.

From Table~\ref{tab:result}, we have three key observations. (1) Our proposed two-stream sequence models and fast forward sequence models achieve significantly better results compared to previous video pooling approaches.  (2) The fast-forward LSTM model with depth 7 can boost the shallow sequence model around 0.5\% in term of GAP. (3) Different temporal modeling approaches are complementary to each other. Our final submission ensembles 57 models with different hidden cells and depths.

\section{Conclusions}
\label{sec:4}

In this work, we have proposed three temporal modeling approaches to address the challenging large-scale video recognition task.
Experiment results verify that our approaches achieve significantly better results than the traditional temporal pooling approaches.
The ensemble of our individual models has been shown to improve the performance further, enabling our method to rank the third place out of 650 teams in the challenge competition. Our PaddlePaddle video toolbox is available for download from \url{https://github.com/baidu/Youtube-8M} and includes implementations of three temporal modeling approaches.


\begin{thebibliography}{10}\itemsep=-1pt

\bibitem{YouTube8M}
S.~Abu-El-Haija, N.~Kothari, J.~Lee, P.~Natsev, G.~Toderici, B.~Varadarajan,
  and S.~Vijayanarasimhan.
\newblock Youtube-8m: A large-scale video classification benchmark.
\newblock {\em arXiv preprint arXiv:1609.08675}, 2016.

\bibitem{arandjelovic2013all}
R.~Arandjelovic and A.~Zisserman.
\newblock All about vlad.
\newblock In {\em CVPR}, pages 1578--1585, 2013.

\bibitem{cho2014properties}
K.~Cho, B.~Van~Merri{\"e}nboer, D.~Bahdanau, and Y.~Bengio.
\newblock On the properties of neural machine translation: Encoder-decoder
  approaches.
\newblock {\em arXiv preprint arXiv:1409.1259}, 2014.

\bibitem{ImageNet}
J.~Deng, W.~Dong, R.~Socher, L.-J. Li, K.~Li, and L.~Fei-Fei.
\newblock Imagenet: A large-scale hierarchical image database.
\newblock In {\em CVPR}, 2009.

\bibitem{gan2016webly}
C.~Gan, C.~Sun, L.~Duan, and B.~Gong.
\newblock Webly-supervised video recognition by mutually voting for relevant
  web images and web video frames.
\newblock In {\em ECCV}, pages 849--866, 2016.

\bibitem{devnet}
C.~Gan, N.~Wang, Y.~Yang, D.-Y. Yeung, and A.~G. Hauptmann.
\newblock Devnet: A deep event network for multimedia event detection and
  evidence recounting.
\newblock In {\em CVPR}, pages 2568--2577, 2015.

\bibitem{gan2016you}
C.~Gan, T.~Yao, K.~Yang, Y.~Yang, and T.~Mei.
\newblock You lead, we exceed: Labor-free video concept learning by jointly
  exploiting web videos and images.
\newblock CVPR, 2016.

\bibitem{resnet}
K.~He, X.~Zhang, S.~Ren, and J.~Sun.
\newblock Deep residual learning for image recognition.
\newblock In {\em Proceedings of the IEEE conference on computer vision and
  pattern recognition}, pages 770--778, 2016.

\bibitem{LSTM}
S.~Hochreiter and J.~Schmidhuber.
\newblock Long short-term memory.
\newblock {\em Neural computation}, 9(8):1735--1780, 1997.

\bibitem{hochreiter1997long}
S.~Hochreiter and J.~Schmidhuber.
\newblock Long short-term memory.
\newblock {\em Neural computation}, 9(8):1735--1780, 1997.

\bibitem{Sports1M}
A.~Karpathy, G.~Toderici, S.~Shetty, T.~Leung, R.~Sukthankar, and L.~Fei-Fei.
\newblock Large-scale video classification with convolutional neural networks.
\newblock In {\em CVPR}, 2014.

\bibitem{HMDB}
H.~Kuehne, H.~Jhuang, E.~Garrote, T.~Poggio, and T.~Serre.
\newblock {HMDB}: a large video database for human motion recognition.
\newblock In {\em ICCV}, pages 2556--2563, 2011.

\bibitem{deep_speaker}
C.~Li, X.~Ma, B.~Jiang, X.~Li, X.~Zhang, X.~Liu, Y.~Cao, A.~Kannan, and Z.~Zhu.
\newblock Deep speaker: an end-to-end neural speaker embedding system.
\newblock {\em arXiv preprint arXiv:1705.02304}, 2017.

\bibitem{ts_lstm}
C.-Y. Ma, M.-H. Chen, Z.~Kira, and G.~AlRegib.
\newblock {TS-LSTM} and temporal-inception: Exploiting spatiotemporal dynamics
  for activity recognition.
\newblock {\em arXiv preprint arXiv:1703.10667}, 2017.

\bibitem{Twostream}
K.~Simonyan and A.~Zisserman.
\newblock Two-stream convolutional networks for action recognition in videos.
\newblock In {\em NIPS}, 2014.

\bibitem{VeryDeep}
K.~Simonyan and A.~Zisserman.
\newblock Very deep convolutional networks for large-scale image recognition.
\newblock {\em ICLR}, 2015.

\bibitem{UCF101}
K.~Soomro, A.~R. Zamir, and M.~Shah.
\newblock {UCF}101: A dataset of 101 human actions classes from videos in the
  wild.
\newblock {\em arXiv preprint arXiv:1212.0402}, 2012.

\bibitem{ULSTM}
N.~Srivastava, E.~Mansimov, and R.~Salakhutdinov.
\newblock Unsupervised learning of video representations using lstms.
\newblock {\em ICML}, 2015.

\bibitem{go_deeper}
C.~Szegedy, W.~Liu, Y.~Jia, P.~Sermanet, S.~Reed, D.~Anguelov, D.~Erhan,
  V.~Vanhoucke, and A.~Rabinovich.
\newblock Going deeper with convolutions.
\newblock In {\em CVPR}, pages 1--9, 2015.

\bibitem{C3D}
D.~Tran, L.~Bourdev, R.~Fergus, L.~Torresani, and M.~Paluri.
\newblock {C3D}: Generic features for video analysis.
\newblock {\em ICCV}, 2015.

\bibitem{xu2015discriminative}
Z.~Xu, Y.~Yang, and A.~G. Hauptmann.
\newblock A discriminative {CNN} video representation for event detection.
\newblock {\em CVPR}, 2015.

\bibitem{zhou2016deep}
J.~Zhou, Y.~Cao, X.~Wang, P.~Li, and W.~Xu.
\newblock Deep recurrent models with fast-forward connections for neural
  machine translation.
\newblock {\em arXiv preprint arXiv:1606.04199}, 2016.

\end{thebibliography}
\end{document}